\pdfoutput=1

\documentclass[11pt]{article}

\usepackage[preprint]{acl}
\usepackage{amsmath}
\usepackage{amssymb}
\usepackage{multirow}

\usepackage{times}
\usepackage{latexsym}
\usepackage{comment}
\usepackage[T1]{fontenc}

\usepackage[utf8]{inputenc}

\usepackage{microtype}

\usepackage{inconsolata}

\usepackage{graphicx}
\usepackage{booktabs}
\usepackage{hyperref}
\usepackage{tabularx}  
\usepackage{enumitem}
\usepackage{float}
\usepackage{makecell}

\newcommand{\llamaThree}{Llama-3-8B-Instruct}
\newcommand{\gemmaThree}{Gemma-3-12B-IT}
\newcommand{\hact}{\mathbf{h}_{\ell, p}}
\newcommand{\hstar}{\mathbf{h}^*}
\newcommand{\hacti}[1]{\mathbf{h}_{\ell, p}^{(#1)}}
\newcommand{\vdir}{\mathbf{v}_{\ell, p}}
\newcommand{\vstar}{\mathbf{v}^{*}}
\newcommand{\vstarnorm}{\hat{\mathbf{v}}^{*}}

\title{Detecting (Un)answerability in Large Language Models\\with Linear Directions}

\author{
 \vspace{7px}
Maor Juliet Lavi ~~~~~~~ Tova Milo ~~~~~~~ Mor Geva \\ \vspace{4px}
Blavatnik School of Computer Science and AI, Tel Aviv University\\
\texttt{\{maorlavi@mail, milo@cs, morgeva@tauex\}.tau.ac.il}
}

\begin{document}
\maketitle
\begin{abstract}

\end{abstract}
Large language models (LLMs) often respond confidently to questions even when they lack the necessary information, leading to hallucinated answers. In this work, we study the problem of (un)answerability detection, focusing on extractive question answering (QA) where the model should determine if a passage contains sufficient information to answer a given question. We propose a simple approach for identifying a direction in the model's activation space that captures unanswerability and uses it for classification. This direction is selected by applying activation additions during inference and measuring their impact on the model's abstention behavior. We show that projecting hidden activations onto this direction yields a reliable score for (un)answerability classification. Experiments on two open-weight LLMs and four extractive QA benchmarks show that our method effectively detects unanswerable questions and generalizes better across datasets than existing prompt-based and classifier-based approaches. 
Moreover, the obtained directions extend beyond extractive QA to unanswerability that stems from factors, such as lack of scientific consensus and subjectivity. Last, causal interventions show that adding or ablating the directions effectively controls the abstention behavior of the model. We release our code at: \url{https://github.com/MaorLavi/unanswerability-directions}.

\section{Introduction}

Large language models (LLMs) often generate confident responses to questions regardless of whether they have the information needed to answer reliably \citep{yin-etal-2023-large,yona-etal-2024-large}. When a model lacks the required information, it often produces inaccurate responses or hallucinations \citep{10.1145/3703155, 05e915a6e62d4a9ba343e39a7189d4f7}, making the identification of such cases an important step toward improving its trustworthiness \citep{kadavath2022languagemodelsmostlyknow, yin-etal-2023-large, amayuelas-etal-2024-knowledge}.
This challenge is particularly important in applications such as medical assistance, legal advice, and educational tools, where incorrect answers can lead to real-world harm. 

\begin{figure}[t]
    \centering
    \includegraphics[scale=0.33]{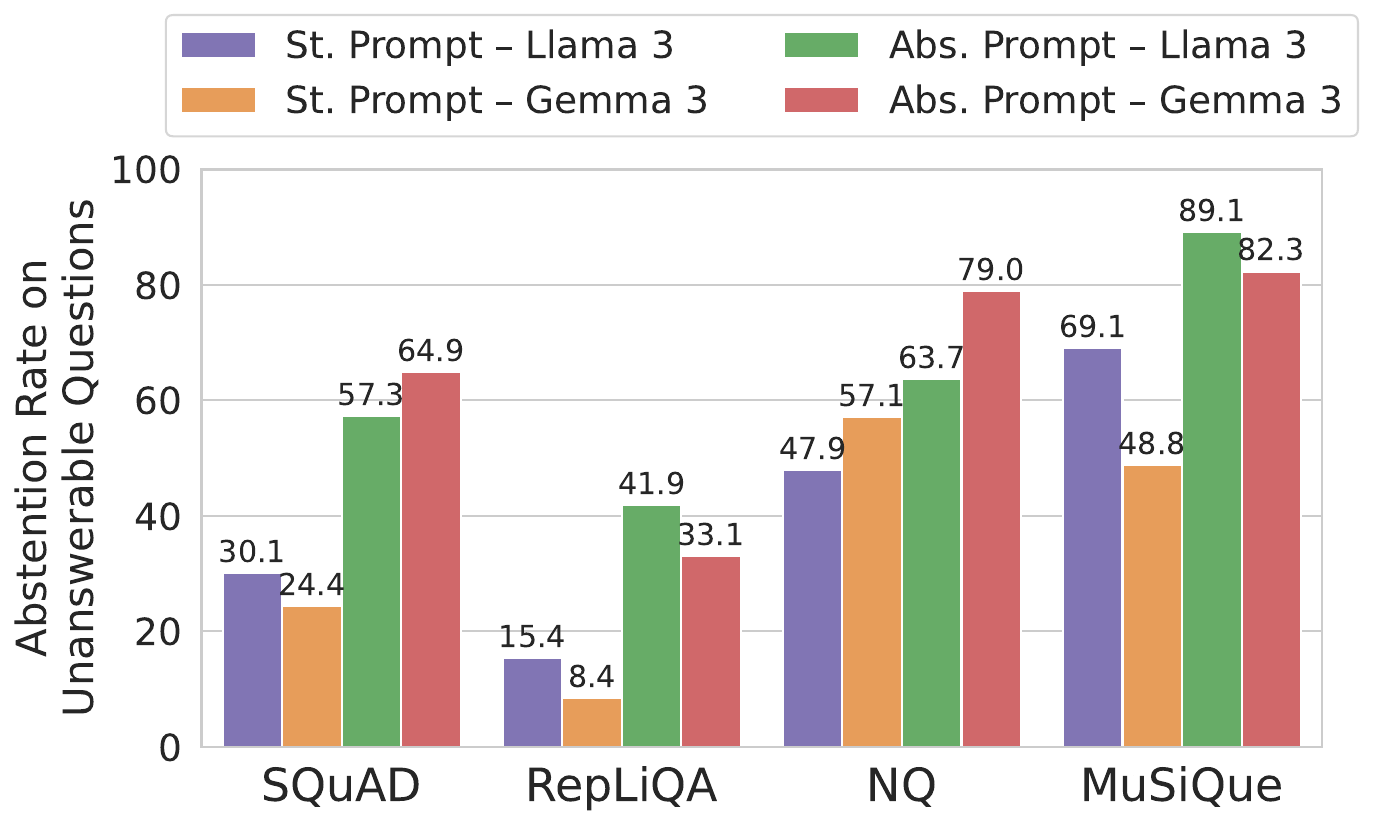}
    \caption{Abstention rate (recall) on unanswerable questions under Standard and Abstain-aware prompts, evaluated on \llamaThree{} and \gemmaThree{}.}
    \label{fig:prompts_recall_unans}
\end{figure}

In this work, we study the problem of unanswerability in the context of extractive question answering (QA), where the model is presented with a question and a passage of text that may or may not contain the information required to answer it \citep{rajpurkar-etal-2018-know}.
As illustrated in Figure~\ref{fig:prompts_recall_unans}, models in this setting tend to respond rather than abstain, even when the question cannot be answered from the provided passage.

Several approaches have been proposed for detecting unanswerable questions. 
Fine-tuning has been suggested to improve abstention behavior in models \citep{feng-etal-2024-dont, zhang-etal-2024-r}.
In extractive QA, prompting has been shown to encourage models to indicate uncertainty \citep{slobodkin-etal-2023-curious}, but performance remains inconsistent across models and datasets.
\citet{slobodkin-etal-2023-curious} further introduced a linear classifier trained on internal model representations to predict unanswerability. Other efforts have explored estimating uncertainty from hidden states \citep{tomani2024uncertaintybasedabstentionllmsimproves, kim2024detectingllmhallucinationlayerwise}, or detecting unanswerable inputs with sparse autoencoder features \citep{heindrich2025do}. While these latter methods have shown promising results, they often fail to generalize across datasets—highlighting a key challenge in robust unanswerability detection.

Here, we analyze the model’s internal activations and show that a single direction in representation space effectively captures unanswerability across diverse datasets. To this end, we first construct a set of candidate directions using  difference-in-means \citep{marks2024geometrytruthemergentlinear}, where the averaged activations of answerable examples are subtracted from those of unanswerable ones at a fixed layer and position. 
To select the most informative direction, we add each candidate vector to the hidden activations at inference time and measure the resulting change in the model’s probability of abstaining.
Finally, we use the selected direction for unanswerability classification:
given an input, we extract its activations at a fixed layer and position and project it onto the learned direction. This projection yields a scalar unanswerability score, which reflects how aligned the model’s internal representation is with unanswerable examples.

We evaluate our method on four question-answering datasets: \textsc{SQuAD} 2.0 \citep{rajpurkar-etal-2016-squad, rajpurkar-etal-2018-know}, \textsc{RepLiQA} \citep{NEURIPS2024_2b236260}, \textsc{Natural Questions} (\textsc{NQ}) \citep{kwiatkowski-etal-2019-natural}, and \textsc{MuSiQue} \citep{trivedi2022musique}, using \llamaThree{} \citep{DBLP:journals/corr/abs-2407-21783} and \gemmaThree{} \citep{gemmateam2025gemma3technicalreport}, and find that the learned direction consistently captures unanswerability.
Our method achieves F1 scores of 75.9--96.4\%, performing comparably to a logistic regression classifier baseline and outperforming prompt-based baselines. Additionally, the direction signal transfers across datasets, exceeding the classifier’s generalization on three out of four datasets by an average of 8.14\%, which is further improved to 9.73\% on average via a simple threshold calibration.
These results highlight the robustness of the learned direction and its ability to generalize across datasets. 
Interestingly, evaluations on SelfAware \citep{yin-etal-2023-large} and CREPE \citep{yu-etal-2023-crepe} show that the directions obtained for the extractive QA datasets further generalize to other settings where unanswerability stems from factors such as subjectivity. 

Last, we validate the signal encoded by the direction through causal interventions, where 
adding the direction vector to the residual stream at a sufficient magnitude causes the model to abstain in nearly all cases (96\%), while ablating it pushes the model to answer even when the context is insufficient.

Beyond classification, our method provides insight into how unanswerability is internally represented by the model, revealing a native signal embedded directly in the representation space. Analysis of failure cases further supports the reliability of this signal. In several instances (26\%), we found that the provided labels were incorrect. Also, in 24\% of cases labeled as answerable, the answer appeared in the passage but not in the context of the specific question, making the instance difficult to classify. A smaller portion (6\%) included questions with grammatical issues, rendering their answerability unclear and dependent on interpretation.

To conclude, we introduce a lightweight and interpretable method for detecting unanswerability in LLMs by uncovering a direction in the model’s activation space that captures an unanswerability signal. We demonstrate the utility of this approach for classifying unanswerable inputs across diverse datasets, and show that the learned direction generalizes better than prompt-based and classifier-based baselines. We also show that we can use this direction to control the model’s tendency to abstain.

\renewcommand{\arraystretch}{1.5}
\begin{table*}[htbp]
\centering
\footnotesize
\setlength{\tabcolsep}{6pt}
\begin{tabular}{@{}p{1.28cm} p{6.7cm} p{4cm} l}
\toprule
\textbf{Dataset} & \textbf{Context} ($c$) & \textbf{Question} ($q$) & \textbf{Label} ($y$) \\
\midrule
\textsc{SQuAD} & In England, the period of Norman architecture immediately succeeds that of the Anglo-Saxon and precedes the Early Gothic... & What architecture type came after Early Gothic? & 1 (unanswerable)\\
\textsc{RepLiQA} & ...One such partnership was formed with the Greenleaf Cafe, a popular downtown eatery, which now organizes 'Saturday Morning Miles'... & What specific event does the Greenleaf Cafe organize as part of Newville's fitness initiative? & 0 (answerable)\\
\textsc{NQ} & The National Professional Soccer League II , which awarded two points for all goals except those on the power play , also used a three - point line... & when did the nba add the three point line ? & 1 (unanswerable)\\
\textsc{MuSiQue} & Ye Rongguang (born October 3, 1963 in Wenzhou, Zhejiang) \dots{} Sanjiang Church was a Christian church located in Yongjia County, near Wenzhou, in Zhejiang Province, China... &  What county was Ye Rongguang born in? & 0 (answerable) \\
\bottomrule
\end{tabular}
\caption{Example context–question pairs from each dataset used in our experiments, labeled as answerable (0) or unanswerable (1).
}
\label{tab:dataset_examples}
\end{table*}

\section{Problem Setup}

We address the task of \textit{unanswerability detection} in extractive QA. Given a context (e.g., passage or document) $c$ and a question $q$, the goal is to determine whether the context contains sufficient information to answer the question. Formally, the input is a pair $(c, q)$, and the objective is to predict a binary label $y \in \{0,1\}$, where $y = 1$ indicates that the question is unanswerable based on the context, and $y = 0$ indicates that it is answerable. 
Examples of answerable and unanswerable cases are shown in Table~\ref{tab:dataset_examples}.

\section{Method}

We take inspiration from prior observations that certain abstract concepts, such as sentiment, refusal, or truthfulness, are linearly encoded within a language model's internal representations \citep[][inter alia]{tigges2023linearrepresentationssentimentlarge, NEURIPS2024_f5454485, marks2024geometrytruthemergentlinear}, 
and aim to identify a direction in the model's activation space that captures unanswerability.
If such a direction exists, it can be used to distinguish answerable from unanswerable instances by measuring the alignment between their internal representations and this direction.
We now describe our methodology for finding such directions in LLMs.

\subsection{Deriving Potential Directions} 
To identify potential directions encoding unanswerability, we follow prior work that uses differences in mean activations between two input sets \citep{marks2024geometrytruthemergentlinear, belrose2024diff, rimsky-etal-2024-steering}. 
Given a model with \( L \) layers and hidden dimension $d$, for each input $(c, q)$ we extract the hidden activations $\hact \in \mathbb{R}^d$ at each layer \( \ell \in \{1, \dots, L\} \) 
and token position \( p \) after the instruction segment.\footnote{These positions correspond to tokens from a chat template 
that wraps chat models' inputs and appear before the model's response, see \S\ref{sec:models} for details}
Let \( \{(c_i, q_i)\}_{i=1}^{N} \) be answerable and \( \{(c_j, q_j)\}_{j=1}^{M} \) unanswerable examples, and let ${\hacti{i}}$ be the hidden activations for the $i$-th input.
We define the candidate direction \( \vdir \in \mathbb{R}^d\) at each layer \( \ell \) and token position \( p \) as the difference between the mean activations over unanswerable and answerable examples:
\[
\vdir = \frac{1}{M} \sum_{j=1}^{M} {\hacti{j}}- \frac{1}{N} \sum_{i=1}^{N} {\hacti{i}}
\]

This yields a set of $L \times N_{\text{pos}}$ directions \( \{\vdir\} \), where \( N_{\text{pos}} \) is the number of token positions considered.

\subsection{Selecting a Direction for Unanswerability}
\label{sec:select_direction}

We employ causal steering \citep{NEURIPS2023_81b83900, DBLP:journals/corr/abs-2308-10248, rimsky-etal-2024-steering} to choose the direction that best represents unanswerability. The selection is done on a separate validation set from the examples used to find the candidate directions.

\paragraph{Activation intervention} For each candidate direction \( \vdir \) and context-question pair \( (c, q) \) in the validation set, we modify the hidden activations 
at the corresponding layer \( \ell \) and position \( p \) as follows:
\[
\tilde{\mathbf{h}}_{\ell, p} = \hact + \vdir.
\]
The modified activations are propagated forward through the model. We repeat this procedure for each candidate direction and analyze its effect on the model’s outputs and abstention behavior.

\paragraph{Steering score} 
\label{sec:steering}
Let \( \{(c_i, q_i)\}_{i=1}^{K} \) denote the validation set, consisting of \( K \) context-question pairs. To approximate abstention behavior, we identify the first token of the word \textit{unanswerable} as it is tokenized by the model (e.g., ``un''), and denote it as \( t_{\text{un}} \in \mathcal{V} \). 
This token is used as a proxy for abstention since the model is prompted to respond with the word \textit{unanswerable} when it cannot answer the question based on the provided context.

For each validation example, we extract the model’s next-token distribution under the intervention.
Let \( p_t^{(i)} \) denote the probability of token \( t \) for the \( i \)-th validation example, the steering score $\psi_{\text{steer}}$
of a direction \( \vdir \) is then defined as:
\[
\psi_{\text{steer}} = \frac{1}{K} \sum_{i=1}^{K} \left[ \log p_{t_{\text{un}}}^{(i)} - \log  \sum_{t \in \mathcal{V} \setminus \{t_{\text{un}}\}} p_t^{(i)}  \right]
\]
This score quantifies how much more likely the model is, on average, to generate \( t_{\text{un}} \) rather than any other token in the vocabulary, when steered with the candidate direction.
Higher values indicate a stronger abstention-inducing effect.

\paragraph{Direction selection} 
We evaluate all \( L \times N_{\text{pos}} \) candidate directions and select the one with the highest steering score. The final unanswerability direction, denoted \(\vstar\), corresponds to the pair \( (\ell^*, p^*) \) that maximizes \(\psi_\text{steer} \).
This selected direction is used in all downstream evaluations and analyses.\footnote{We also tested whether combining two directions from different layers and token positions improves performance, and find that it does not yield substantial gains (see \S\ref{appendix:multilayer}).}

\subsection{Unanswerability Classification}
We use the selected direction $\vstar$ to define a scalar scoring function that quantifies how strongly a given input aligns with the unanswerability signal; this score is then used to classify new inputs as answerable or unanswerable.

\paragraph{Unanswerability score}
Let \( \vstarnorm\) denote the normalized direction selected in the previous step.
For a given context-question pair \( (c, q) \), we extract the hidden activations \(\hstar\in \mathbb{R}^d \) from the selected layer $\ell^*$ and position $p^*$. The unanswerability score is computed as the dot product between this hidden state and the normalized direction:
\begin{align*}
\phi_{\text{unans}} = \langle \hstar, \vstarnorm \rangle
\end{align*}

This scalar is intended to reflect how strongly the input aligns with the learned unanswerability signal. Since it is unbounded and varies across models and datasets, we next describe how we interpret this value for classification.

\paragraph{Thresholding the unanswerability score}
\label{sec:thresholding}
To establish a classifier, we select the threshold \( \tau \) on the unanswerability score using the validation set. Specifically, we compute the ROC curve and choose \( \tau \) to minimize the Euclidean distance to the ideal point (TPR = 1, FPR = 0).
At inference time, for a given input $(c,q)$, 
if $\phi_\text{unans}$ exceeds \( \tau \), then the input is classified as unanswerable; otherwise, it is classified as answerable.

\section{Experiments}
We evaluate our method against three baselines and report classification accuracy and generalization across datasets, as well as a causal analysis of the learned direction.
Our results show that: (1) the direction-based method achieves strong performance when derived and evaluated on the same dataset, close to a trained classifier and outperforming prompt-based baselines; (2) the direction generalizes more robustly across datasets than the classifier, especially after a lightweight threshold calibration; and (3) the selected directions causally influence the model’s abstention behavior.

\subsection{Experimental Setup}

\paragraph{Datasets} 
We evaluate our method on four question answering benchmarks---\textsc{SQuAD} 2.0~\citep{rajpurkar-etal-2018-know}, \textsc{RepLiQA}~\citep{NEURIPS2024_2b236260}, \textsc{NQ}~\citep{kwiatkowski-etal-2019-natural}, and \textsc{MuSiQue}~\citep{trivedi2022musique}---all structured as context-question pairs.

\textsc{SQuAD} 2.0 and \textsc{RepLiQA} natively include explicitly labeled answerable and unanswerable examples.
\textsc{SQuAD} 2.0 augments the original \textsc{SQuAD} dataset, which is based on Wikipedia articles, with unanswerable questions that appear plausible given the context. \textsc{RepLiQA} is constructed from human-written reference documents across diverse topics not found on the web, so models cannot rely on their parametric knowledge.

For \textsc{NQ} and \textsc{MuSiQue}, we use the versions of the datasets curated by \citet{slobodkin-etal-2023-curious}. \textsc{NQ} consists of real user questions paired with Wikipedia paragraphs, while \textsc{MuSiQue} contains multi-hop questions created by composing seed questions from various datasets. The curated version retains the original answerable examples and constructs unanswerable ones by replacing gold paragraphs with semantically similar ones that do not answer the question.
(see \S\ref{appendix:experimental_details} for more details).

For each dataset, we sample a total of 4,000 examples, which we split into training (1,200), development (800), and test (2,000) sets, with an equal number of answerable and unanswerable instances in each split. Table~\ref{tab:dataset_examples} provides representative examples from each dataset.

\paragraph{Models}
\label{sec:models}
We experiment with two instruction-tuned models: \llamaThree{} \citep{DBLP:journals/corr/abs-2407-21783} and \gemmaThree{} \citep{gemmateam2025gemma3technicalreport}. Both were trained with chat templates that wrap the user instruction (see \S\ref{appendix:chat_templates} for the full templates). In our analysis, we focus on hidden activations at the positions of the template tokens that immediately follow the user instruction, 
as they represent the model’s internal state after processing the full context and question and just before it begins generating a response. In addition, all inputs are formatted using the Abstain-aware Prompt (see \S\ref{sec:baselines}). 

\begin{figure*}[htbp]
    \centering
    \includegraphics[scale=0.45]{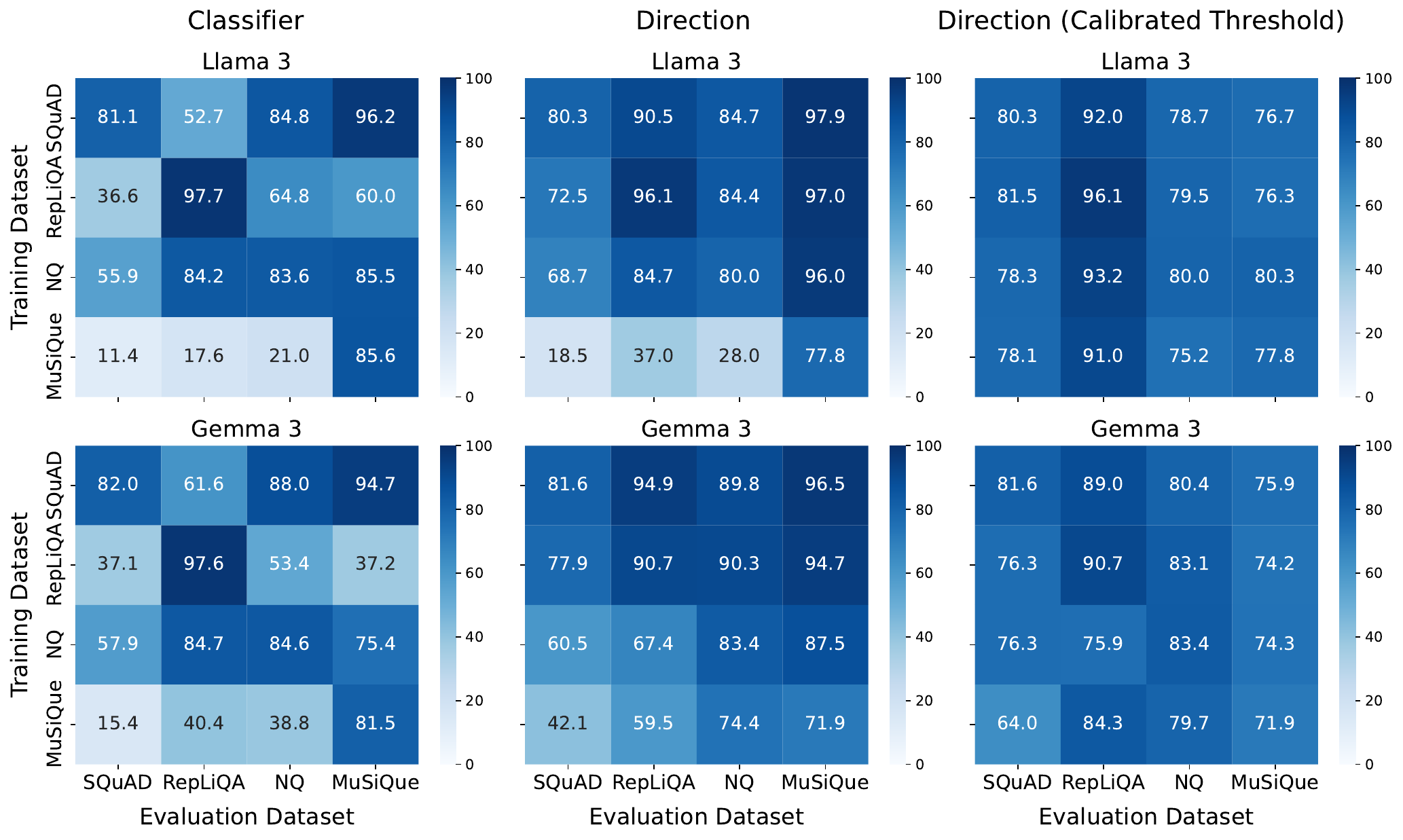}
    \caption{Unanswerable prompts recall (abstention rate) across datasets using three methods: a trained classifier, a direction-based method with a fixed threshold, and a calibrated threshold variant. Each heatmap shows generalization performance from training on one dataset (rows) to evaluating on another (columns). Results are shown for both \llamaThree{} (top) and \gemmaThree{} (bottom). See \S\ref{appendix:statistics} for statistical tests comparing the methods.
    }
    \label{fig:unans_recall_heatmap}
\end{figure*}

\paragraph{Method}
We find that the token ``un'' corresponds to the first token in \textit{unanswerable} in both \llamaThree{} and \gemmaThree{}, and set it as \(t_{\text{un}}\). We apply the method described in \S\ref{sec:steering} to select the layer and token position for each model–dataset pair, and find that the selected layers consistently lie near the middle of the model. This is consistent with prior work suggesting that middle layers in transformer models tend to capture abstract semantic properties, in contrast to lower layers which focus on lexical patterns and upper layers which are more task-specific \citep{tenney-etal-2019-bert, jawahar-etal-2019-bert, vulic-etal-2020-probing, geva-etal-2021-transformer}. Classification thresholds are set using ROC curves on the validation sets (see \S\ref{appendix:dir_select} for direction and threshold selection details).

\paragraph{Baselines} 
\label{sec:baselines}

We compare our method against the following baselines: 
\begin{itemize}
  \item \textit{Standard Prompt}: A prompt-only baseline where the model is given the context and question without any additional instruction.

  \item \textit{Abstain-aware Prompt}: A prompt augmentation baseline, in which an instruction is added encouraging the model to abstain if the question is unanswerable \citep{slobodkin-etal-2023-curious}.

  \item \textit{Classifier}: A logistic regression model trained on hidden activations to predict unanswerability \citep{slobodkin-etal-2023-curious}. The classifier is trained using cross-validation on the combined training and validation sets, with model inputs formatted using the Abstain-aware Prompt.
\end{itemize}
Full prompt templates for the prompt-based baselines are provided in \S\ref{appendix:baseline_prompts}.

\paragraph{Evaluation metrics}
We measure precision, recall, and F1 score separately for the answerable and unanswerable classes. 
We also report macro-average F1 score, which balances precision and recall across both classes equally. Since the prompt-based baselines generate textual output, we first classify each response as either an abstention or an attempt to answer the question. 
To do so, we use GPT-4o mini \citep{openai2024gpt4omini}, prompted with instructions and few-shot examples to make this decision. The full prompt used and a manual analysis validating this automatic evaluation are in \S\ref{appendix:gpt_prompt}.

\begin{figure*}[ht]
    \centering
    \includegraphics[scale=0.45]{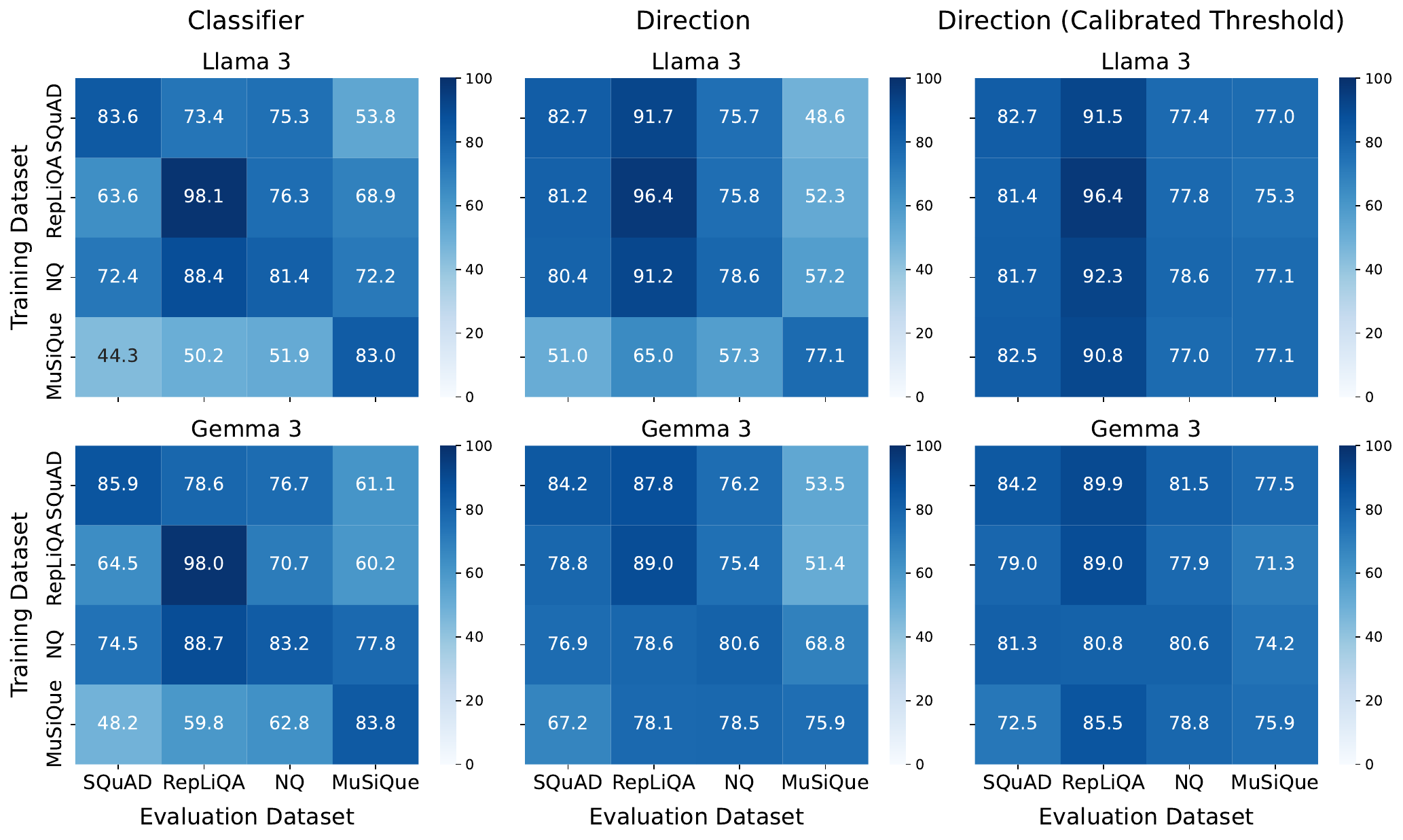}
    \caption{Macro-average F1 scores across datasets using three methods: a trained classifier, a direction-based method with a fixed threshold, and a calibrated threshold variant. Each heatmap shows generalization performance from training on one dataset (rows) to evaluating on another (columns). Results are shown for both \llamaThree{} (top) and \gemmaThree{} (bottom). See \S\ref{appendix:statistics} for statistical tests comparing the methods.}
    \label{fig:f1_heatmap}
\end{figure*}

\subsection{(Un)answerability Classification}
We evaluate how effectively our method distinguishes answerable and unanswerable questions.

\paragraph{Direction-based method effectively detects unanswerable questions} 
Figure~\ref{fig:unans_recall_heatmap} (left and middle) shows the recall on unanswerable examples for our direction-based method and the classifier baseline, across all combinations of evaluation and source datasets.
Figure~\ref{fig:prompts_recall_unans} shows the recall on unanswerable examples for the Standard Prompt and the Abstain-aware Prompt baselines. Both the classifier and our method outperform the prompt-based baselines. When the training and test splits are from the same dataset, the classifier achieves the highest overall recall, averaging 87\% for \llamaThree{} and 86.4\% for \gemmaThree{}. Our direction-based method is slightly below, with an average recall of 83.6\% and 81.9\%, respectively. However, when evaluated on unseen datasets, the classifier performance drops by an average of 30.2\%, while our method drops by only 7.4\%, demonstrating better generalization.

\begin{figure}[ht]
    \centering
    \includegraphics[scale=0.33]{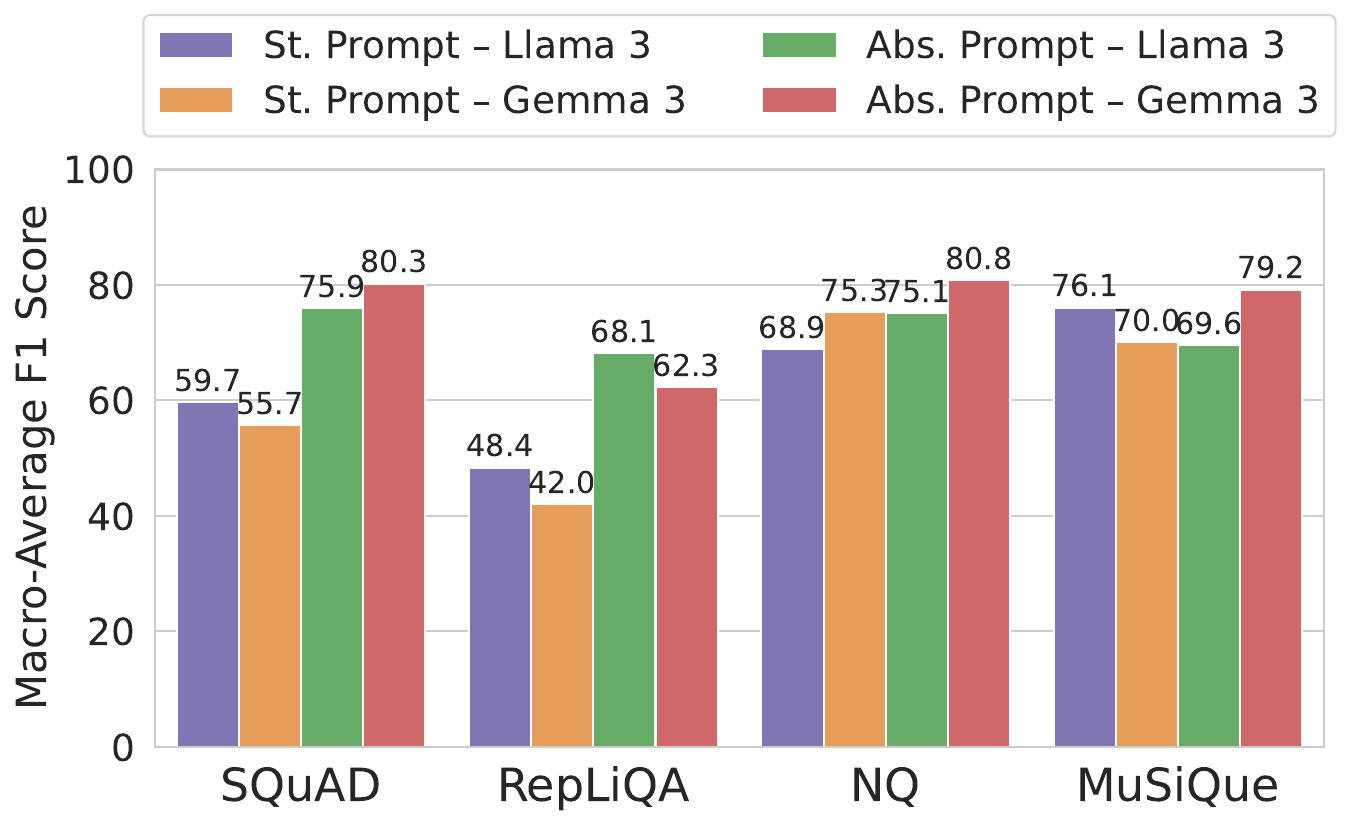}
    \caption{Macro-average F1 scores on answerable and unanswerable questions under Standard and Abstain-aware prompts, evaluated on \llamaThree{} and \gemmaThree{}.}
    \label{fig:prompts_f1}
\end{figure}

\paragraph{Direction-based classification outperforms baselines on unseen datasets} 
Figure~\ref{fig:f1_heatmap} (left and middle) presents the macro-average F1 scores for our method and the classifier baseline across all source–evaluation dataset pairs, and Figure~\ref{fig:prompts_f1}  
reports scores for the prompt-based baselines. We observe the same trends in F1 scores: when the direction is evaluated on the same dataset is was derived from, it achieves 83.7\% on average on \llamaThree{} and 82.4\% on \gemmaThree{}, compared to 86.5\% and 87.8\%, respectively, for the classifier.
However, our method demonstrates stronger generalization than the classifier baseline when evaluated on unseen datasets. Specifically, on \textsc{SQuAD}, \textsc{RepLiQA}, and \textsc{NQ}, it outperforms the classifier by 1.8--11.9\%, averaged per evaluation dataset.
The only exception is \textsc{MuSiQue}, where the classifier generalizes better by 8.4--12.2\%. 
We will next show that these results can be improved with a simple threshold calibration, indicating that even in cases where the direction appears not to generalize well, the issue lies in the decision boundary rather than in the quality of the signal itself.
 
\paragraph{Threshold calibration further improves generalization}
To understand whether the weaker generalization results reflect that the direction captured a dataset-specific signal, or simply a need for threshold calibration, we visualize the unanswerability scores \( \phi_{\text{unans}} \) across datasets (see \S\ref{appendix:projection_scores}).
We observe that the direction consistently induces a separation between answerable and unanswerable examples,
however, the optimal decision threshold varies between datasets. This motivates refining the threshold using the validation set of each evaluation dataset, without modifying the learned direction itself, following the procedure described in \S\ref{sec:thresholding}.

As shown in Figures~\ref{fig:unans_recall_heatmap} and \ref{fig:f1_heatmap}, with dataset-specific thresholding, the direction-based method achieves consistent performance across evaluation datasets, regardless of its source. This simple calibration improves generalization results by 2.7--23.7\% across evaluation datasets, achieving performance only 2.6\% lower on average than that of directions derived from the same datasets.
These results suggests that the unanswerability signal captured by the direction is robust and consistently encoded across datasets.

\subsection{Steering Effectiveness}
\begin{figure}[t]
    \centering
    \includegraphics[scale=0.33]{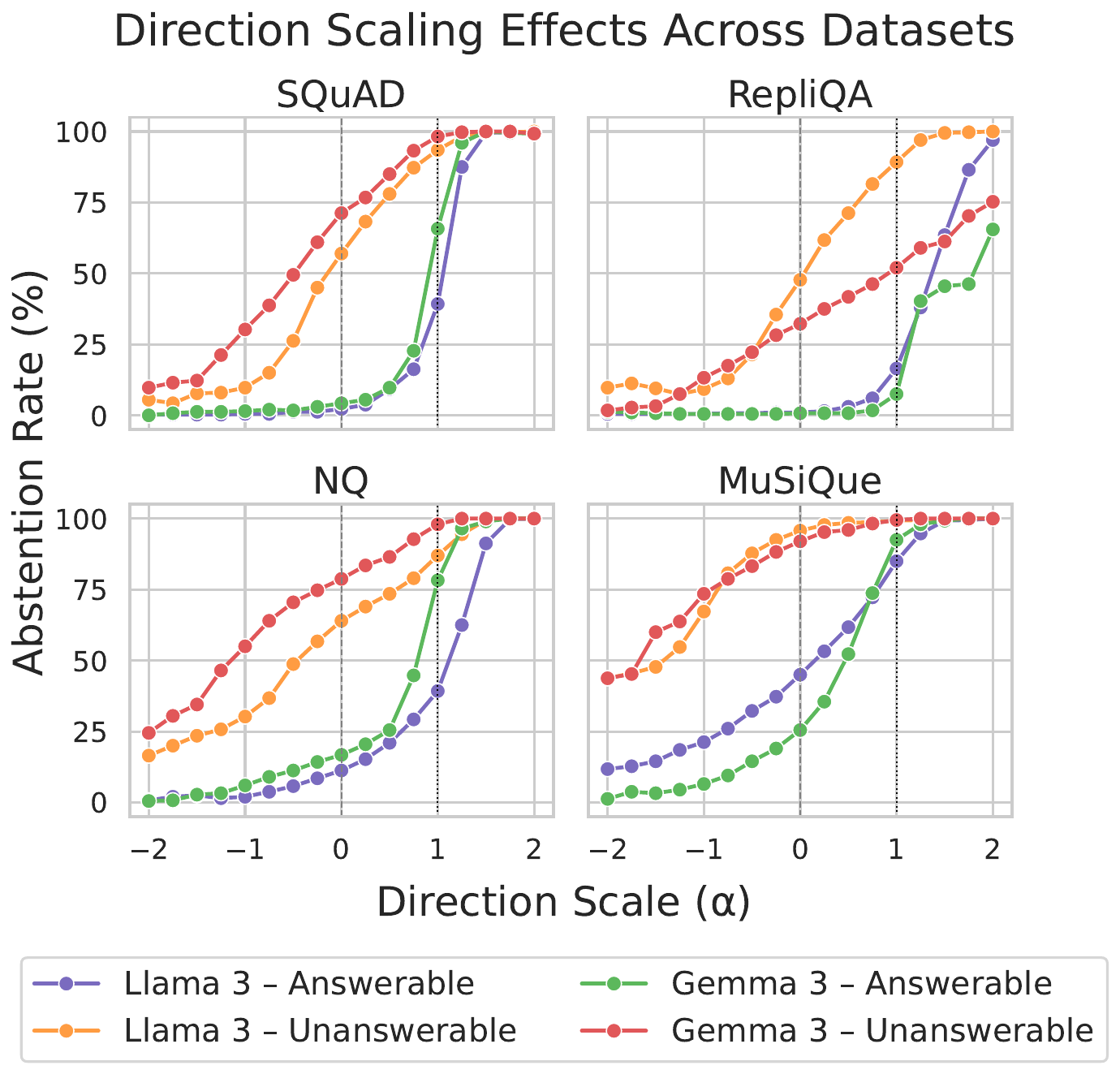}
    \caption{Effect of activation interventions on model abstention rates across steering strengths ($\alpha$). Results are shown for both answerable and unanswerable validation examples, for each dataset and model. 
    }
    \label{fig:causal_interventions}
\end{figure}

To further show that the selected direction captures an unanswerability signal and to observe whether it can influence abstention, we assess its causal impact. To do so, We perform activation space interventions at the chosen layer \(\ell^*\) and token position \(p^*\), for each dataset and model. For a given context--question pair \((c, q)\) formatted with the Abstain-aware Prompt, we modify the hidden activations at layer \(\ell^*\) and position \(p^*\) by adding the selected direction, normalized scaled by \(\alpha\):
\begin{align*}
\tilde{\mathbf{h}}^* = \hstar + \alpha\, \vstarnorm
\end{align*}
where \(\alpha \in [-2, 2]\) controls the strength and polarity of the intervention. We use GPT-4o mini to determine if the model abstained or attempted to answer the question (see \S\ref{appendix:gpt_prompt}), and measure the abstention rate on both answerable and unanswerable validation examples under each intervention (see Figure~\ref{fig:causal_interventions}). 
In all cases, increasing \(\alpha\) leads to a sharp rise in abstention on both unanswerable and answerable inputs, with mean abstention rates (across all datasets) reaching 96.8\% and 95.2\%, respectively, at \(\alpha = 2.0\). Conversely, when \(\alpha = -2.0\), abstention drops to 2.0\% for answerable prompts and 19.4\% for unanswerable ones. These results provide strong evidence that the direction causally influences the model’s decision to abstain.

\subsection{Generalization Beyond Extractive QA}
We further evaluate our method, using the directions we had already derived from the extractive QA datasets, on SelfAware \citep{yin-etal-2023-large} and CREPE \citep{yu-etal-2023-crepe}. SelfAware contains both answerable and unanswerable questions, where unanswerability arises from lack of scientific consensus, imagination, subjectivity, too many variables, and philosophical questions. CREPE consists of open-domain questions with false presuppositions, collected from online forums.

Results are presented in Tables~\ref{tab:selfaware} and~\ref{tab:crepe}, showing that with threshold calibration our method achieves F1 scores in the ranges of 71.3\%--78.67\% for SelfAware (Table~\ref{tab:selfaware}) and 59.12\%--61.85\% for CREPE (Table~\ref{tab:crepe}). Overall, this shows that the directions captured for unanswerability in extractive QA settings generalize to other settings, transferring well to SelfAware, although the performance on CREPE is lower.

\begin{table}[t]
\centering
\small
\begin{tabular}{lcc}
\toprule
\makecell{Training\\Dataset} & \makecell{Direction\\w/o calibration} & \makecell{Direction \\ w calibration} \\
\midrule
SQuAD   & 56.67 & 74.46 \\
RepLiQA & 54.14 & 78.67 \\
NQ      & 56.74 & 74.25 \\
MuSiQue & 46.03 & 71.30 \\
\bottomrule
\end{tabular}
\caption{Evaluation of directions obtained from different extractive QA datasets on the SelfAware dataset (F1 scores).}
\label{tab:selfaware}
\end{table}

\begin{table}[t]
\centering
\small
\begin{tabular}{lcc}
\toprule
\makecell{Training\\Dataset} & \makecell{Direction \\ w/o calibration} & \makecell{Direction \\ w/ calibration} \\
\midrule
SQuAD   & 45.97 & 61.85 \\
RepLiQA & 36.94 & 61.23 \\
NQ      & 42.74 & 60.37 \\
MuSiQue & 35.56 & 59.12 \\
\bottomrule
\end{tabular}
\caption{Evaluation of directions  obtained from different extractive QA datasets on the CREPE dataset (F1 scores).}
\label{tab:crepe}
\end{table}

\section{Error Analysis}
\label{sec:analysis}
To better understand the limitations of our method, we conducted a manual categorization of 100 misclassified examples: 50 from \textsc{SQuAD} and 50 from \textsc{NQ}, evenly split between answerable and unanswerable instances. Each was assigned to one of five categories: 
\begin{itemize}
    \item \textit{Direction Failure}: the direction score led to an incorrect prediction despite a correct label and well-formed input.
    \item \textit{Incorrect Label}: the ground-truth annotation appears wrong.
    \item \textit{Required Title}: (\textsc{SQuAD} only) the document title (not included in our inputs) was necessary to interpret the passage.
    \item \textit{Grammar ``Mistake''}: ungrammatical phrasing or ambiguity made the input difficult to interpret.
    \item \textit{Answer Not in Context}: the answer exists in the passage but is not clearly in the context of the question.
\end{itemize}
Table~\ref{tab:failure_analysis} shows the results. We find that 53\% of the errors are due to direction failures, and 26\% stem from annotation errors, especially among \textsc{NQ} \textit{unanswerable} examples. Notably, 24\% of the misclassified \textit{answerable} examples fall into the “answer not in context” category, most of them in \textsc{NQ}. Overall, this categorization reveals that many of the model’s errors arise from ambiguous inputs or limitations in the dataset, rather than clear failures of the method itself.
Representative examples from each category are included in \S\ref{tab:failure_examples}.

\begin{table}[t]
\centering
\small
\begin{tabular}{@{}p{1.6cm} ccccc@{}}
\toprule
\multirow{2}{*}{\textbf{Category}} & \multicolumn{2}{c}{\textbf{\textsc{SQuAD}}} & \multicolumn{2}{c}{\textbf{\textsc{NQ}}} & \multirow{2}{*}{\begin{tabular}{@{}c@{}}\textbf{Overall} \\ \%\end{tabular}}
 \\
 & \textbf{Ans} & \textbf{Unans} & \textbf{Ans} & \textbf{Unans} & \\
\midrule
Direction failure               & 14 & 20 & 7  & 12 & 53 \\
Incorrect $\;$ label                & 5  & 3  & 7  & 11 & 26 \\
Required $\quad$ title                 & 3  & 0  & 0  & 0  & 3  \\
Grammar ``mistake''            & 0  & 2  & 2  & 2  & 6 \\
Answer not in context          & 3  & 0  & 9  & 0  & 12 \\
\bottomrule
\end{tabular}
\caption{Manual categorization of 100 direction-based classification errors, evenly sampled from \textsc{SQuAD} and \textsc{NQ} (with 25 answerable and 25 unanswerable examples from each).}
\label{tab:failure_analysis}
\end{table}

\section{Related Work}

Prior work has explored methods to improve abstention behavior in models: \citet{Lan2020ALBERT:} improved reasoning with a pretraining loss, leading to improved performance on QA tasks, including unanswerable questions, whereas \citet{Zhang_Yang_Zhao_2021} introduced a verification process to detect when questions cannot be answered. Fine-tuned approaches have also been proposed to reduce hallucinations by improving the model’s ability to abstain \citep{zhang-etal-2024-r, feng-etal-2024-dont}. In contrast, we detect unanswerability by interpreting internal representations of the model, leaving it unchanged.

Several works \citep{tomani2024uncertaintybasedabstentionllmsimproves, kim2024detectingllmhallucinationlayerwise} evaluated model uncertainty as a signal for whether a question could be answered given the context. We, however, focus directly on unanswerability detection, without estimating uncertainty.

Prompt manipulations were also proposed to detect unanswerability, but showed unstable performance across datasets and models \citep{slobodkin-etal-2023-curious, zhou-etal-2023-context}. \citet{slobodkin-etal-2023-curious} further identified an unanswerability-related subspace by training a logistic regression classifier on last-layer hidden representations. Here, we aim to identify a direction in activation space that influences the model's abstention behavior and captures unanswerability consistently across datasets.
Another approach used sparse autoencoder features to classify unanswerable inputs \citep{heindrich2025do}. Though effective on the training dataset, the generalization ability of the last two methods proved inconsistent. In contrast, our approach offers a lightweight method for unanswerability classification and demonstrates stronger generalization across datasets.

Extracting linear directions from model activation has been a common technique for analyzing and modifying model behavior \citep{NIPS2016_a486cd07, NEURIPS2023_81b83900, marks2024geometrytruthemergentlinear, hong2025reasoningmemorizationinterplaylanguagemodels, cohen2025pretrainedllmslearnmultiple}.
In this work, we show that similar techniques can be applied to identify a direction associated with unanswerability, and demonstrate how we can use this direction to classify whether a question can be answered from the given context.

\section{Conclusion}
Our work introduces a method for identifying a direction in the model's activation space that captures unanswerability, using difference-in-means and a selection criterion based on activation steering. We introduce a simple classification method that uses this direction to detect unanswerable questions. We compare our method to existing approaches and find that, while the strongest baseline achieves slightly higher performance when evaluated on its training dataset, our method generalizes more effectively across datasets. We also show that causal interventions along the direction induce abstention behavior of the model. These findings support the view that abstract properties such as unanswerability are linearly encoded in the intermediate representations of language models, and show that this signal can be leveraged for both interpretation and practical use.

\section*{Limitations}
Our approach assumes that unanswerability is mediated by a linear direction from a fixed layer and token position. While we capture a strong signal and observe that combining two directions does not improve performance, it is possible that unanswerability is represented in more complex patterns that our method cannot identify.
In addition, we use a simple threshold over the projection onto the direction for classification and do not explore more expressive functions, which could potentially better exploit this signal. 
Finally, while our experiments show that the method generalizes beyond extractive QA to some extent, there may be more effective ways to capture a broader unanswerability signal that transfers robustly across diverse settings.

\section*{Acknowledgments}
This work was supported in part by the Alon scholarship, the Intel Rising Star Faculty Award, the Israel Science Foundation (ISF) grant 1083/24, and the ISF grant 2707/22 of the Breakthrough Research Grant (BRG) Program.

\bibliography{custom}

\appendix

\section{Experimental Setup - Additional Details}
This section provides additional details about our experimental setup, including further details on the curated datasets and prompt templates used in our experiments
\label{appendix:experimental_details}

\paragraph{Curated Versions of \textsc{NQ} and \textsc{MuSiQue}}
We use the curated versions of \textsc{NQ} and \textsc{MuSiQue} introduced by \citet{slobodkin-etal-2023-curious}. In \textsc{NQ}, each example consists of a real user question paired with a paragraph from a Wikipedia article. Answerable instances are drawn from questions that include both a long and short answer; the long answer is used as context. Unanswerable instances are constructed by replacing the context with a semantically similar paragraph from the same article that is \textit{not} annotated as the long answer. Paragraphs are ranked using cosine similarity over Sentence-BERT embeddings.

\textsc{MuSiQue} is a multi-hop QA benchmark in which each instance includes a complex question, a decomposition into sub-questions, and a set of candidate paragraphs. In the curated version, answerable examples are formed by concatenating the gold paragraphs aligned with each sub-question. To generate unanswerable examples, one or more of these gold paragraphs are replaced with the most semantically similar but incorrect paragraphs, identified using the same retrieval method applied in \textsc{NQ}.

\paragraph{Model Chat Templates}
\label{appendix:chat_templates}

\llamaThree{} and \gemmaThree{} are instruction-tuned using system-defined chat templates that wrap the user instruction before response generation.we use these same templates in our experiments, as shown in Table~\ref{tab:chat_templates}. As described in \S\ref{sec:models}, we extract hidden activations at the template positions following the user instruction.

\begin{table}[t]
\centering
\footnotesize
\renewcommand{\arraystretch}{1.2}
\begin{tabular}{p{2cm} c}
\toprule
\textbf{Model} & \textbf{Chat Template} \\
\midrule
\multirow{4}{2cm}{\llamaThree{}} & \texttt{<|start\_header\_id|>user} \\
& \texttt{<|end\_header\_id|>\{instruction\}} \\
& \texttt{<|eot\_id|><|start\_header\_id|>} \\
& \texttt{assistant<|end\_header\_id|>} \\
\midrule
\multirow{3}{2cm}\gemmaThree{} & 
\texttt{<start\_of\_turn>user} \\
& \texttt{\{instruction\}<end\_of\_turn>} \\
& \texttt{<start\_of\_turn>model} \\
\bottomrule
\end{tabular}
\caption{Chat templates used to format the user instruction during inference.}
\label{tab:chat_templates}
\end{table}

\paragraph{Prompt-based Baseline Prompts}
\label{appendix:baseline_prompts}
Table~\ref{tab:default_prompt} shows the prompt used in the standard prompt-based baseline, which contains only the context and question. Table~\ref{tab:abstention_instruction_prompt} presents the modified version used in the abstention-instruction baseline, which encourages the model to abstain when the question cannot be answered from the passage.

\begin{table}[htbp]
\footnotesize
\centering
\begin{tabularx}{0.8\columnwidth}{|X|}
\toprule 
Given the following passage and question, answer the question. \\
Passage: <passage> \\
Question: <question> \\
Answer: \\
\bottomrule
\end{tabularx}
\caption{Prompt used in the standard prompt-based baseline.}
\label{tab:default_prompt}
\end{table}

\begin{table}[htbp]
\footnotesize
\centering
\begin{tabularx}{0.8\columnwidth}{|X|}
\toprule 
Given the following passage and question, answer the question. \\
First make sure if it can be answered by the passage. \\
If it cannot be answered based on the passage, reply "unanswerable". \\
Passage: <passage> \\
Question: <question> \\
Answer: \\
\bottomrule
\end{tabularx}
\caption{Prompt used in the abstention-instruction baseline.}
\label{tab:abstention_instruction_prompt}
\end{table}

\paragraph{Evaluating Prompt-Based Baselines with GPT-4o mini}
\label{appendix:gpt_prompt}
Table~\ref{tab:gpt4o_prompt} displays the full prompt given to GPT-4o-mini to determine whether a model’s response constitutes an abstention. The prompt includes detailed instructions and few-shot examples. To assess the reliability of this evaluation method, we conducted a manual evaluation over 50 model responses: 25 express abstention and 25 attempt to answer. The responses were sampled from model outputs generated for inputs from our datasets and labeled manually. GPT-4o-mini correctly classified all 50 examples.

\begin{table*}[htbp]
\footnotesize
\begin{tabularx}{\textwidth}{|X|}
\toprule 
You will be shown a model's response to a question. Your task is to decide whether the model attempted to answer the question. \\
It is considered to abstain if it explicitly says it cannot answer the question, indicates uncertainty, or refuses to provide an answer because the context is insufficient. \\
Otherwise, it is considered to have answered the question. \\

Respond with "yes" if the model attempted to answer the question. \\
Respond with "no" if the model abstained. \\
You will respond with a single word: "yes" or "no". Do not provide any explanation or extra text. \\\\

\textbf{Examples:} \\
Model response: "1598" \\
Answer: yes \\\\
Model response: "abc" \\
Answer: yes \\\\
Model response: "The answer can be answered by the passage. According to the passage, Fort Coligny was destroyed in 1560." \\
Answer: yes \\\\
Model response: "According to the passage, problems capable of theoretical solutions but consuming unreasonable time in practical application are known as intractable problems" \\
Answer: yes \\\\
Model response: "The passage does not mention BSkyB agreeing with any claim. In fact, it states that BSkyB denied Virgin Media's claim that they had substantially increased the asking price for the channels. Therefore, the answer is \"unanswerable\"." \\
Answer: no \\\\
Model response: "The passage does not answer the question. The passage only mentions that the five counties (Los Angeles, Orange, San Diego, San Bernardino, and Riverside) are among the top 15 most populous counties in the United States, but it does not provide the actual names of the five most populous counties in the United States." \\
Answer: no \\\\
Model response: "The passage does not mention the birth year of Reguier de la Plancha. It only mentions that he died in 1560, but does not provide his birth year." \\
Answer: no \\\\
\bottomrule
\end{tabularx}
\caption{Prompt used to instruct GPT-4o mini to classify model responses as either abstentions or answer attempts.}
\label{tab:gpt4o_prompt}
\end{table*}
\clearpage
\paragraph{Direction and Threshold Selection}
\label{appendix:dir_select}
Table~\ref{tab:selected_dirs} shows the layer and token position selected for each model–dataset pair, based on the method described in \S\ref{sec:select_direction}. For \llamaThree{}, the same layer and position (layer 16, position $-1$) were selected across all datasets. For \gemmaThree{}, the selected layers range from 26 to 27, with positions $-1$ or $-4$, depending on the dataset. Since \llamaThree{} has 32 layers and \gemmaThree{} has 48, the selected layers lie near the middle of each model.

\begin{figure}[htb]
    \centering
    \includegraphics[scale=0.4]{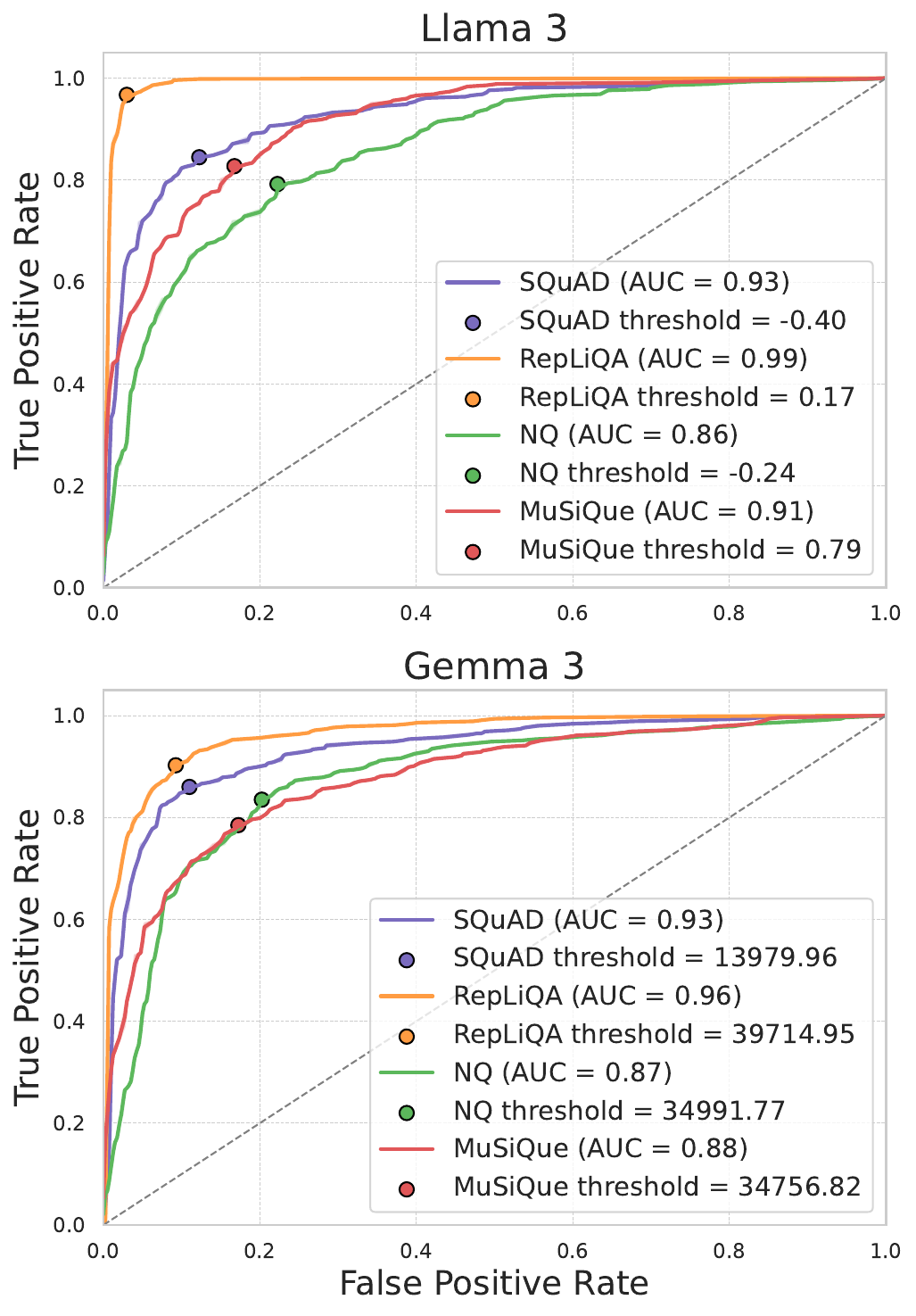}
    \caption{ROC curves across datasets using \llamaThree{} and \gemmaThree{}. The x-axis shows the false positive rate (answerable instances classified as unanswerable), and the y-axis shows the true positive rate (unanswerable instances correctly identified). Each curve is annotated with its AUC, and markers indicate the selected classification threshold per dataset.}
    \label{fig:roc_curves}
\end{figure}

We compute ROC curves on the validation sets (Figure~\ref{fig:roc_curves}) to select classification thresholds based on the separation between answerable and unanswerable examples. The chosen threshold minimizes the distance to the ideal point and is applied at test time.

\begin{table}[htb]
\centering
\small
\begin{tabular}{p{1.6cm} p{2.23cm} c c }

\toprule
\textbf{Model} & \textbf{Dataset} & \textbf{Layer} & \textbf{Position} \\
\midrule
\llamaThree{}     & All              & 16             & $-1$              \\
\midrule
\multirow{4}{1.8cm}{\gemmaThree{}}
  & \textsc{SQuAD}             & 27 & $-1$ \\
  & \textsc{RepLiQA}           & 26 & $-1$ \\
  & \textsc{NQ} & 27 & $-1$ \\
  & \textsc{MuSiQue}           & 27 & $-4$ \\
\bottomrule
\end{tabular}
\caption{Selected layer and token position from which hidden activations were extracted to compute the unanswerability direction for each model–dataset pair. 
}
\label{tab:selected_dirs}
\end{table}

\section{Full Classification Results}
\label{appendix:full_results}

We report the full classification results in Tables~\ref{tab:full_metrics_squad}, \ref{tab:full_metrics_repliqa}, \ref{tab:full_metrics_nq}, and \ref{tab:full_metrics_musique}. For each direction, derived from \textsc{SQuAD}, \textsc{RepLiQA}, \textsc{NQ}, and \textsc{MuSiQue}, we present precision, recall, and F1 scores for both the \textit{answerable} and \textit{unanswerable} classes, across all evaluation datasets and for both \llamaThree{} and \gemmaThree{}. Results are reported both with the original threshold and after applying threshold calibration.

\begin{table*}[htb]
\centering
\footnotesize
\setlength{\tabcolsep}{6pt}
\renewcommand{\arraystretch}{1}
\begin{tabular}{llcccccccccccc}
\toprule
\multirow{3}{*}{\textbf{Eval Dataset}} & \multirow{3}{*}{\textbf{Class}} &
\multicolumn{6}{c}{\textbf{Llama 3}} & \multicolumn{6}{c}{\textbf{Gemma 3}} \\
& &
\multicolumn{3}{c}{\textbf{Original Threshold}} &
\multicolumn{3}{c}{\textbf{+ Calibration}} &
\multicolumn{3}{c}{\textbf{Original Threshold}} &
\multicolumn{3}{c}{\textbf{+ Calibration}} \\
& & \textbf{P} & \textbf{R} & \textbf{F1} & \textbf{P} & \textbf{R} & \textbf{F1} & \textbf{P} & \textbf{R} & \textbf{F1} & \textbf{P} & \textbf{R} & \textbf{F1} \\
\midrule
\multirow{2}{*}{\textsc{SQuAD}} & Ans   & 81.2 & 85.2 & 83.2 & -- & -- & -- & 82.5 & 86.8 & 84.6 & -- & -- & -- \\
                                & Unans & 84.4 & 80.3 & 82.3 & -- & -- & -- & 86.1 & 81.6 & 83.8 & -- & -- & -- \\
\midrule
\multirow{2}{*}{\textsc{RepLiQA}} & Ans   & 90.7 & 92.9 & 91.8 & 91.9 & 90.9 & 91.4 & 94.1 & 80.8 & 86.9 & 89.2 & 90.8 & 90.0 \\
                                  & Unans & 92.7 & 90.5 & 91.6 & 91.0 & 92.0 & 91.5 & 83.2 & 94.9 & 88.7 & 90.6 & 89.0 & 89.8 \\
\midrule
\multirow{2}{*}{\textsc{NQ}} & Ans   & 81.4 & 67.1 & 73.6 & 78.1 & 76.1 & 77.1 & 86.1 & 63.4 & 73.0 & 80.8 & 82.5 & 81.6 \\
                             & Unans & 72.0 & 84.7 & 77.9 & 76.7 & 78.7 & 77.7 & 71.0 & 89.8 & 79.3 & 82.1 & 80.4 & 81.3 \\
\midrule
\multirow{2}{*}{\textsc{MuSiQue}} & Ans   & 88.7 & 16.4 & 27.7 & 76.8 & 77.2 & 77.0 & 86.8 & 23.0 & 36.4 & 76.7 & 79.1 & 77.9 \\
                                  & Unans & 53.9 & 97.9 & 69.6 & 77.1 & 76.7 & 76.9 & 55.6 & 96.5 & 70.6 & 78.4 & 75.9 & 77.1 \\
\bottomrule
\end{tabular}
\caption{Full classification results using the direction derived from \textsc{SQuAD}. For each evaluation dataset and class, we report precision (P), recall (R), and F1 score for \llamaThree{} and \gemmaThree{}, under the original threshold and after applying threshold calibration.}
\label{tab:full_metrics_squad}
\end{table*}

\begin{table*}[htb]
\centering
\footnotesize
\setlength{\tabcolsep}{6pt}
\renewcommand{\arraystretch}{1}
\begin{tabular}{llcccccccccccc}
\toprule
\multirow{3}{*}{\textbf{Eval Dataset}} & \multirow{3}{*}{\textbf{Class}} &
\multicolumn{6}{c}{\textbf{Llama 3}} & \multicolumn{6}{c}{\textbf{Gemma 3}} \\
& &
\multicolumn{3}{c}{\textbf{Original Threshold}} &
\multicolumn{3}{c}{\textbf{+ Calibration}} &
\multicolumn{3}{c}{\textbf{Original Threshold}} &
\multicolumn{3}{c}{\textbf{+ Calibration}} \\
& & \textbf{P} & \textbf{R} & \textbf{F1} & \textbf{P} & \textbf{R} & \textbf{F1} & \textbf{P} & \textbf{R} & \textbf{F1} & \textbf{P} & \textbf{R} & \textbf{F1} \\
\midrule
\multirow{2}{*}{\textsc{SQuAD}} & Ans   & 76.6 & 90.1 & 82.8 & 81.5 & 81.3 & 81.4 & 78.3 & 79.7 & 79.0 & 77.5 & 81.7 & 79.6 \\
                                & Unans & 88.0 & 72.5 & 79.5 & 81.3 & 81.5 & 81.4 & 79.3 & 77.9 & 78.6 & 80.7 & 76.3 & 78.4 \\
\midrule
\multirow{2}{*}{\textsc{RepLiQA}} & Ans   & 96.1 & 96.7 & 96.4 & -- & -- & -- & 90.4 & 87.4 & 88.9 & -- & -- & -- \\
                                  & Unans & 96.7 & 96.1 & 96.4 & -- & -- & -- & 87.8 & 90.7 & 89.2 & -- & -- & -- \\
\midrule
\multirow{2}{*}{\textsc{NQ}} & Ans   & 81.3 & 67.6 & 73.8 & 78.8 & 76.2 & 77.5 & 86.4 & 61.5 & 71.9 & 81.2 & 72.9 & 76.8 \\
                             & Unans & 72.3 & 84.4 & 77.9 & 77.0 & 79.5 & 78.2 & 70.1 & 90.3 & 78.9 & 75.4 & 83.1 & 79.1 \\
\midrule
\multirow{2}{*}{\textsc{MuSiQue}} & Ans   & 87.7 & 21.3 & 34.3 & 75.8 & 74.3 & 75.1 & 80.0 & 21.2 & 33.5 & 72.6 & 68.4 & 70.4 \\
                                  & Unans & 55.2 & 97.0 & 70.4 & 74.8 & 76.3 & 75.5 & 54.6 & 94.7 & 69.3 & 70.1 & 74.2 & 72.1 \\
\bottomrule
\end{tabular}
\caption{Full classification results using the direction derived from \textsc{RepLiQA}. For each evaluation dataset and class, we report precision (P), recall (R), and F1 score for \llamaThree{} and \gemmaThree{}, under the original threshold and after applying threshold calibration.}
\label{tab:full_metrics_repliqa}
\end{table*}

\begin{table*}[htb]
\centering
\footnotesize
\setlength{\tabcolsep}{6pt}
\renewcommand{\arraystretch}{1}
\begin{tabular}{llcccccccccccc}
\toprule
\multirow{3}{*}{\textbf{Eval Dataset}} & \multirow{3}{*}{\textbf{Class}} &
\multicolumn{6}{c}{\textbf{Llama 3}} & \multicolumn{6}{c}{\textbf{Gemma 3}} \\
& &
\multicolumn{3}{c}{\textbf{Original Threshold}} &
\multicolumn{3}{c}{\textbf{+ Calibration}} &
\multicolumn{3}{c}{\textbf{Original Threshold}} &
\multicolumn{3}{c}{\textbf{+ Calibration}} \\
& & \textbf{P} & \textbf{R} & \textbf{F1} & \textbf{P} & \textbf{R} & \textbf{F1} & \textbf{P} & \textbf{R} & \textbf{F1} & \textbf{P} & \textbf{R} & \textbf{F1} \\
\midrule
\multirow{2}{*}{\textsc{SQuAD}} & Ans   & 74.7 & 92.6 & 82.7 & 74.2 & 93.3 & 82.6 & 70.5 & 94.6 & 80.8 & 78.5 & 86.5 & 82.3 \\
                                & Unans & 90.3 & 68.7 & 78.0 & 91.0 & 67.5 & 77.5 & 91.8 & 60.5 & 72.9 & 85.0 & 76.3 & 80.4 \\
\midrule
\multirow{2}{*}{\textsc{RepLiQA}} & Ans   & 86.5 & 97.8 & 91.8 & 85.4 & 97.9 & 91.2 & 73.5 & 90.3 & 81.0 & 78.1 & 85.8 & 81.8 \\
                                  & Unans & 97.5 & 84.7 & 90.6 & 97.5 & 83.3 & 89.9 & 87.4 & 67.4 & 76.1 & 84.2 & 75.9 & 79.9 \\
\midrule
\multirow{2}{*}{\textsc{NQ}} & Ans   & 79.5 & 77.3 & 78.4 & -- & -- & -- & 82.4 & 77.9 & 80.1 & -- & -- & -- \\
                             & Unans & 77.9 & 80.0 & 78.9 & -- & -- & -- & 79.1 & 83.4 & 81.2 & -- & -- & -- \\
\midrule
\multirow{2}{*}{\textsc{MuSiQue}} & Ans   & 87.6 & 28.2 & 42.7 & 87.5 & 29.3 & 43.9 & 80.6 & 52.0 & 63.2 & 74.2 & 74.0 & 74.1 \\
                                  & Unans & 57.2 & 96.0 & 71.7 & 57.5 & 95.8 & 71.9 & 64.6 & 87.5 & 74.3 & 74.1 & 74.3 & 74.2 \\
\bottomrule
\end{tabular}
\caption{Full classification results using the direction derived from \textsc{NQ}. For each evaluation dataset and class, we report precision (P), recall (R), and F1 score for \llamaThree{} and \gemmaThree{}, under the original threshold and after applying threshold calibration.}
\label{tab:full_metrics_nq}
\end{table*}

\begin{table*}[htb]
\centering
\footnotesize
\setlength{\tabcolsep}{6pt}
\renewcommand{\arraystretch}{1}
\begin{tabular}{llcccccccccccc}
\toprule
\multirow{3}{*}{\textbf{Eval Dataset}} & \multirow{3}{*}{\textbf{Class}} &
\multicolumn{6}{c}{\textbf{Llama 3}} & \multicolumn{6}{c}{\textbf{Gemma 3}} \\
& &
\multicolumn{3}{c}{\textbf{Original Threshold}} &
\multicolumn{3}{c}{\textbf{+ Calibration}} &
\multicolumn{3}{c}{\textbf{Original Threshold}} &
\multicolumn{3}{c}{\textbf{+ Calibration}} \\
& & \textbf{P} & \textbf{R} & \textbf{F1} & \textbf{P} & \textbf{R} & \textbf{F1} & \textbf{P} & \textbf{R} & \textbf{F1} & \textbf{P} & \textbf{R} & \textbf{F1} \\
\midrule
\multirow{2}{*}{\textsc{SQuAD}} & Ans   & 55.0 & 99.6 & 70.9 & 79.9 & 87.0 & 83.3 & 62.7 & 97.3 & 76.3 & 69.3 & 81.4 & 74.9 \\
                                & Unans & 97.9 & 18.5 & 31.1 & 85.7 & 78.1 & 81.7 & 94.0 & 42.1 & 58.2 & 77.5 & 64.0 & 70.1 \\
\midrule
\multirow{2}{*}{\textsc{RepLiQA}} & Ans   & 61.3 & 99.9 & 76.0 & 91.0 & 90.7 & 90.8 & 70.8 & 98.4 & 82.4 & 84.7 & 86.8 & 85.7 \\
                                  & Unans & 99.7 & 37.0 & 54.0 & 90.7 & 91.0 & 90.9 & 97.4 & 59.5 & 73.9 & 86.5 & 84.3 & 85.4 \\
\midrule
\multirow{2}{*}{\textsc{NQ}} & Ans   & 57.3 & 96.7 & 72.0 & 76.1 & 78.8 & 77.4 & 76.4 & 82.7 & 79.4 & 79.4 & 78.0 & 78.7 \\
                             & Unans & 89.5 & 28.0 & 42.7 & 78.0 & 75.2 & 76.6 & 81.1 & 74.4 & 77.6 & 78.4 & 79.7 & 79.0 \\
\midrule
\multirow{2}{*}{\textsc{MuSiQue}} & Ans   & 77.5 & 76.5 & 77.0 & -- & -- & -- & 74.0 & 80.0 & 76.9 & -- & -- & -- \\
                                  & Unans & 76.8 & 77.8 & 77.3 & -- & -- & -- & 78.2 & 71.9 & 74.9 & -- & -- & -- \\
\bottomrule
\end{tabular}
\caption{Full classification results using the direction derived from \textsc{MuSiQue}. For each evaluation dataset and class, we report precision (P), recall (R), and F1 score for \llamaThree{} and \gemmaThree{}, under the original threshold and after applying threshold calibration}
\label{tab:full_metrics_musique}
\end{table*}

\paragraph{Statistical Validation of Results}
\label{appendix:statistics}
We conducted statistical tests comparing the performance of our method to that of the linear classifier across 12 training–evaluation dataset combinations of generalization. Specifically, we followed the guidelines of \citet{dror-etal-2018-hitchhikers} and applied a permutation test (10,000 iterations) and McNemar’s test to evaluate whether our method achieves significantly better accuracy ($p < 0.05$).  

With threshold calibration, the direction significantly outperformed the classifier in 91.67\% (11/12) of combinations on \llamaThree{} and 83.33\% (10/12) on \gemmaThree{}, on both the permutation and McNemar’s tests (Table~\ref{tab:stats_cal}). Without threshold calibration, the direction outperformed the classifier in 58.33\% (7/12) of combinations for both models under the permutation test, and in 50\% (6/12) for \llamaThree{} and 58.33\% (7/12) for \gemmaThree{} under McNemar’s test (Table~\ref{tab:stats_nocal}). Overall, these results show that our method finds directions that often outperform the classifier, with consistent significant improvements after calibration.

\section{Combining Directions Across Layers and Token Positions}
\label{appendix:multilayer}
We conducted an experiment testing whether combining two directions from different layers and token positions improves performance. Specifically, we selected the top two vectors (based on maximum steering scores), taken from different layers and token positions, and for classification calculated the two corresponding dot products and summed the scores. Table \ref{tab:two_vectors} shows the averaged F1 scores across all dataset combinations for both our original method and the two-vector method, as well as the average difference. Overall, we see that combining two directions doesn't lead to substantial gains in performance.

\clearpage
\begin{table}[htb]
\centering
\small
\begin{tabular}{p{2.5cm}cc}
\toprule
Model & \makecell{Permutation \\ test} & \makecell{McNemar’s \\ test} \\
\midrule
\llamaThree{} & \makecell{91.67\% \\ (11/12)} & \makecell{91.67\% \\ (11/12)} \\
\gemmaThree{} & \makecell{83.33\% \\ (10/12)} & \makecell{83.33\% \\ (10/12)} \\
\bottomrule
\end{tabular}
\caption{Statistical significance of direction vs.\ classifier with threshold calibration.}
\label{tab:stats_cal}
\end{table}

\begin{table}[htb]
\centering
\small
\begin{tabular}{p{2.5cm}cc}
\toprule
Model & \makecell{Permutation \\ test} & \makecell{McNemar’s \\ test} \\
\midrule
\llamaThree{} & \makecell{58.33\% \\ (7/12)} & \makecell{50\% \\ (6/12)} \\
\gemmaThree{} & \makecell{58.33\% \\ (7/12)} & \makecell{58.33\% \\ (7/12)} \\
\bottomrule
\end{tabular}
\caption{Statistical significance of direction vs.\ classifier without threshold calibration.}
\label{tab:stats_nocal}
\end{table}

\begin{table}[htb]
\centering
\small
\setlength{\tabcolsep}{1pt}
\begin{tabular}{lcc}
\toprule
 & \makecell{Direction \\ (original threshold)} & \makecell{Direction \\ (+ calibration)} \\
\midrule
\makecell{Avg.\ F1 \\ (1 vector)}   & 72.64 & 82.30 \\
\makecell{Avg.\ F1 \\ (2 vectors)}  & 73.27 & 82.49 \\
\makecell{Avg.\ diff.\ (\%)}        & 0.63  & 0.19 \\
\makecell{SD\ (\%)} & 4.04  & 0.61 \\
\bottomrule
\end{tabular}
\caption{Comparison of classification performance using one vs.\ two directions from different layers and token positions.}
\label{tab:two_vectors}
\end{table}

\section{Unanswerability Score Distributions}
\label{appendix:projection_scores}

\begin{figure}[htb]
    \centering
    \includegraphics[scale=0.32]{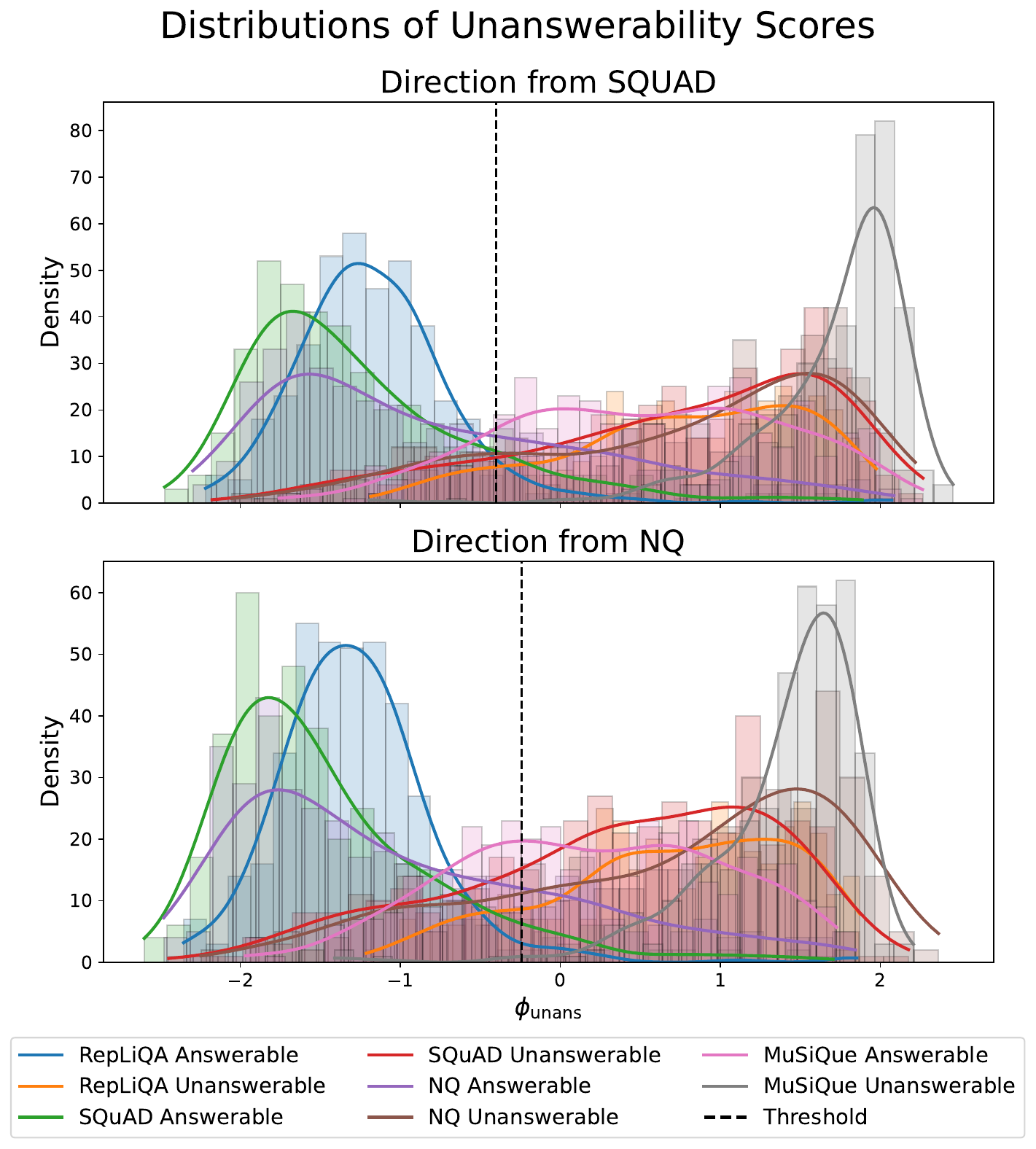}
    \caption{Distribution of unanswerability scores $\phi_{\text{unans}}$ across datasets using the directions derived from \textsc{SQuAD} and \textsc{NQ} in \llamaThree{}.}
    \label{fig:llama_project_scores}
\end{figure}

To visualize how well the unanswerability direction separates answerable and unanswerable examples, we plot the unanswerability scores $\phi_\text{unans}$, i.e the results of the projection of hidden activations onto a direction derived from a specific dataset. For each target dataset, we display the distribution of scores for both classes using a fixed direction. Figure~\ref{fig:llama_project_scores} show the distributions for \llamaThree{} using directions derived from \textsc{SQuAD} and \textsc{NQ}, respectively. In both cases, the direction induces clear separation between the answerable and unanswerable classes.
However, the score distributions vary across datasets, suggesting that the optimal decision threshold differs depending on the evaluation set.
\section{Example Failure Cases}
\label{tab:failure_examples}

Table~\ref{tab:failure_case_examples} provides one representative example for each of the error categories described in \S\ref{sec:analysis}. Each row includes the input question, context, predicted label, and a brief explanation of the failure.

\section{Resources and Packages}
\label{sec:resources}

In our experiments, we used models and data from the transformers \cite{wolf-etal-2020-transformers} and datasets \citep{lhoest-etal-2021-datasets} packages. AI models (specifically ChatGPT) were used to implement certain helper functions. All the experiments were conducted using a single H100 80GB GPU.

\begin{table*}[t]
\centering
\footnotesize
\setlength{\tabcolsep}{4pt}
\begin{tabular}{p{2.2cm} p{1cm} p{1.7cm} p{4cm} p{2cm} p{3cm}}
\toprule
\textbf{Category} & \textbf{Dataset} & \textbf{Label} & \textbf{Context} & \textbf{Question} & \textbf{Comment} \\
\midrule

Direction Failure & \textsc{SQuAD} & Unanswerable & 
The topic of language for writers from Dalmatia and Dubrovnik... These facts undermine the Croatian language proponents' argument that modern-day Croatian is based on a language called Old Croatian. & 
Prior to the 19th century where did Croatians and Serbians live? & 
The direction predicts “answerable,” but the context passage does not answer the question. \\

\midrule

Incorrect Label & \textsc{NQ} & Unanswerable & 
The ileum is the third and final part of the small intestine... It ends at the ileocecal junction... & 
Where is the ileum located in the body? & 
Based on the passage, the ileum is located in the small intestine, specifically as the third and final part of it. Therefore, the label is incorrect. \\

\midrule

Required Title & \textsc{SQuAD} & Answerable & 
During the latter half of the 20th century, a more diverse range of industry also came to the city, including aircraft and car manufacture... & 
Southampton's range of industries includes the manufacture of cars and what other transport? & 
Without the passage title (“Southampton”), it is unclear that “the city” refers to the correct subject. \\

\midrule

Grammar $\quad$ ``Mistake'' & \textsc{SQuAD} & Unanswerable & 
...MCA agent Lew Wasserman made a deal with Universal for his client James Stewart... & 
Who was a MAC agent? & 
The question refers to “MAC” while the passage only mentions “MCA” so the correct label is unanswerable. However, the model may have treated this as a minor typo, leading the direction to misclassify it as answerable. \\

\midrule

Answer Not $\quad \;$ in Context & \textsc{NQ} & Answerable & 
The Speaker, Majority Leader, Minority Leader, Majority Whip and Minority Whip all receive special office suites... & 
Top 5 leadership positions in the House of Representatives? & 
The roles are listed, but the passage does not clearly frame them as leadership positions relevant to the question. \\

\bottomrule
\end{tabular}
\caption{Representative misclassified examples from each failure category. Each includes the dataset, gold label, and a brief explanation of the error.}
\label{tab:failure_case_examples}
\end{table*}

\end{document}